\documentclass[10pt,journal,compsoc]{IEEEtran}
\ifCLASSOPTIONcompsoc
  \usepackage[nocompress]{cite}
\else
  \usepackage{cite}
\fi
\ifCLASSINFOpdf
\else
\fi

\usepackage{amssymb}
\usepackage{extarrows}
\usepackage{graphicx}
\usepackage{subfigure}
\usepackage{algorithm}
\usepackage{algorithmic}
\usepackage{amsmath,bm,url}
\usepackage{cite}
\usepackage{booktabs}
\usepackage[bookmarksnumbered,bookmarksopen,bookmarks=true,unicode=true]{hyperref}
\usepackage{adjustbox}
\usepackage{graphicx}

\hypersetup{hidelinks}
\begin{document}
\title{Dynamic State Warping}

\author{Zhichen Gong
       and Huanhuan Chen,~\IEEEmembership{Senior Member,~IEEE}
        % <-this % stops a space
\IEEEcompsocitemizethanks{\IEEEcompsocthanksitem Z. Gong and H. Chen are with UBRI, School of Computer Science and
Technology, University of Science and Technology of China, Hefei 230027,
China (e-mail: zcgong@mail.ustc.edu.cn; hchen@ustc.edu.cn;}
% <-this % stops an unwanted space
}

\IEEEtitleabstractindextext{%
\begin{abstract}
The ubiquity of sequences in many domains enhances significant recent
interest in sequence learning, for which a basic problem is how to measure the distance between
sequences.
Dynamic time warping (DTW) aligns two sequences by nonlinear local warping and
returns a distance value. DTW shows superior ability in many applications,
e.g. video, image, etc.
However, in DTW, two points are paired
essentially based on point-to-point Euclidean distance (ED) without considering
the autocorrelation of sequences.
Thus, points with different semantic meanings, e.g. peaks and valleys, may be matched providing their coordinate
values are similar.
As a result, DTW is sensitive to noise and poorly interpretable.
This paper proposes an efficient and flexible sequence
alignment algorithm, dynamic state warping (DSW). DSW converts each time
point into a latent state, which endows point-wise autocorrelation
information. Alignment is performed by using the state sequences. Thus DSW
is able to yield alignment that is semantically more
interpretable than that of DTW.
%By aligning the representations of sequences we take into account the history and current input time point
%of sequences.
%In particular, we use a recurrent neural network to learn the representations.
%Our method can deal with uni-variate and multi-variate sequences naturally
%without adjust the algorithm. %using the same framework.
Using one nearest neighbor classifier,
DSW shows significant improvement on classification
accuracy in comparison to ED (70/85 wins) and DTW (74/85
wins). We also empirically demonstrate that DSW is more robust and scales
better to long sequences than ED and DTW.
%comparative methods.
%The implementation of DSW is available online\footnote{%
%www.github.com}.
%%The ubiquity of sequences in many domains enhances signi?cant recent interest in sequence data mining. How to measure the distance between sequences is a basic problem in many applications. Dynamic time war ping (DTW) is to align two sequences by allowing nonlinear local war ping and returns a distance value. DTW shows superior ability in many applications, such as video, image, etc. However, in DTW, two points are paired essentially based on point-to-point Euclidean distance without taking into account the autocorrelation information of sequences. Thus, points with different semantic meanings, e.g. peaks and valleys, may be matched providing their coordinate values are similar. As a result, DTW is sensitive to noise and poorly interpretable. As a remedy, we propose an ef?cient and ?exible sequence alignment algorithm, dynamic state war ping (DSW). DSW converts each time point into a latent state, which endows point-wise autocorrelation structure information. Alignment is performed by using the state sequences. Thus DSW is able to yield alignment between sequences that is semantically more interpretable than that of DTW. DSW is evaluated on synthetic and benchmark sequences. Using one nearest neighbor classi?er, DSW shows signi?cant improvement on classi?cation accuracy in comparison to Euclidean distance (70/85 wins) and DTW (74/85 wins). We also empirically demonstrate that DSW is more robust and scales better to long sequences than Euclidean distance and DTW.
\end{abstract}

}

\maketitle

\IEEEdisplaynontitleabstractindextext

% \fi
\IEEEpeerreviewmaketitle

\IEEEraisesectionheading{\section{Introduction}
\label{section_introduction}}

\IEEEPARstart{S}{equences} are generated and analyzed in almost every domain of human society
such as medical \cite{jo2015time}, engineering \cite{chen2014learning},
entertainment \cite{goroshin2015unsupervised}, etc.
%Significant research interests have been given to time series.
Computing the distance between sequences is critical for classification and
has attracted significant research interest \cite{pei2016modeling,chen2013model,chen2015model}.

For sequence classification, one nearest neighbor (1NN) classifier has been
empirically shown to be a strong solution with proper distance measurements \cite%
{bagnall2016great,ye2009time,rakthanmanon2012searching}. 1NN classifier is
intrinsically parameter-free. In this case, the only concern is how to
measure the distance between sequences properly \cite{batista2014cid}.

Note that sequences are different from typical vectorial data. Sequences are
high dimensional, auto-correlated among temporal axes and possibly of varying
length. Therefore, to measure the distance between sequences, consideration
has to be given to the properties of sequences, such as nonlinear local
warping, phase shift and scaling distortion etc. A more comprehensive review
can be found in \cite{batista2014cid}.

Dynamic time warping (DTW) \cite{berndt1994using,keogh2001derivative} is to
compute the distance between two sequences by warping them locally to the
same length. It allows one-to-many mappings between sequences to ``stretch''
a sequence or many-to-one mappings to ``condense'' a sequence. In this way,
DTW is naturally compatible with phase distortion invariance of sequential
data \cite{batista2014cid}. Despite the simplicity of DTW, 1NN
classifier with DTW distance has been very successful in many applications,
such as video, image and audio etc. \cite%
{zhou2016generalized,rakthanmanon2012searching,kogan1998automated}. It has
been a consensus that DTW may be the strongest distance measurements for
sequences \cite{ding2008querying,begum2015accelerating}.

\begin{figure}[!t]
\centering
\includegraphics[width=3.3in]{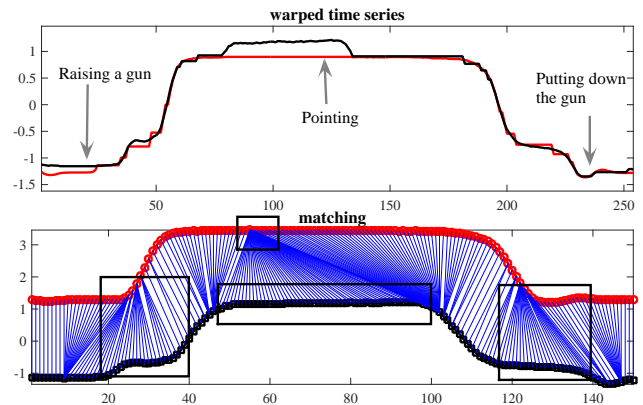}
\caption{DTW alignment of GunPoint dataset. The sequences presents three
stages clearly, i.e. raising a gun, pointing and putting down the gun. DTW
is unable to find the discriminative features between two sequences ([20$\sim
$40] and [120$\sim$140]). We also observe a severe distortion that DTW
matches one point on the upside sequence to almost the whole subsequence of
the downside sequence.}
\label{fig_gundtwmatch}
\end{figure}

%However, DTW still has some limitations.
DTW aligns two sequences by taking the point-to-point comparison of
coordinate values as a fundamental unit \cite{begum2015accelerating}.
Concretely, one point on a sequence is compared to all or a subset of points
on the other sequence to compute a point level distance. In doing so, one
can find an alignment path such that the aligned sequences yields a globally
minimum Euclidean distance \cite{berndt1994using}, which is returned as the
distance between the original sequences. However, this point-level
comparison usually cannot provide dynamic evidence for matching two
points. Note that from human's intuition, we intend to pair two points of
two sequences by taking into account the nearby points or even the global
structure. Thus, DTW does align globally but is unable to take into
consideration the auto-correlated structure information properly \cite%
{ye2009time}.
%In particular, DTW finds the optimal alignment path that has the globally minimum accumulative distance.
%In particular, no local structure characteristics has been taken into consideration.
%However, DTW takes the Euclidean distance between two time points as the only evidence for matching them, which
%limits it from taking advantage of abundant local information.
This makes DTW fragile to noise \cite{keogh2001derivative}, which is also demonstrated in our
experiment (Section \ref{section_Experiment}). As pointed
out in \cite{ye2009time,ye2011time}, DTW usually has weak
interpretability of its alignment. Concretely, the alignment result may lack
local semantic meanings. For example, DTW may match points on a local peak
and a valley if their Euclidean distance is small. Besides, DTW may match
one point to too many points resulting in un-intuitive alignment results and
degrade the classification performance \cite{chen2014learning}.

\begin{figure}[!t]
\centering
\includegraphics[width=3.3in]{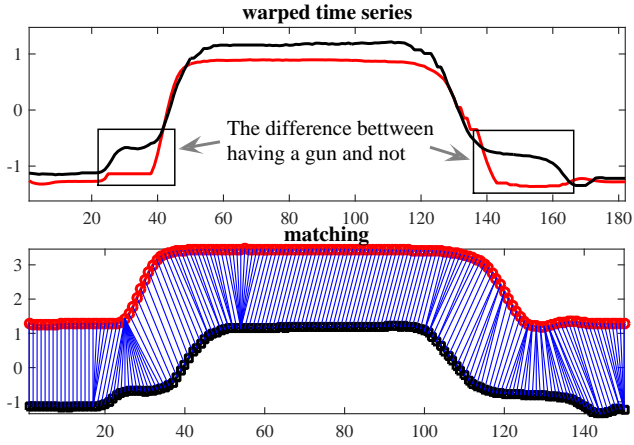}
\caption{DSW alignment of GunPoint dataset. The alignment of DSW is
semantically more sensible than that of DTW. The discriminative features,
i.e. having a gun or not, is detected.}
\label{fig_gundswmatch}
\end{figure}

Figure \ref{fig_gundtwmatch} illustrates the DTW match result of GunPoint
dataset from UCR time series archive \cite{UCRArchive}. The two sequences
are from two different classes. Briefly speaking, GunPoint dataset contains
two class of motion track sequences of an actress. Each motion includes
three processes: raising a gun, pointing and putting down the gun. For the
first class, the gun is initially in a hip-mounted holster. For the second
class, the gun is initially in the hand of the actress instead of the
holster. In time points roughly ranging in [20$\sim$ 40] and [120$\sim$ 140], the
two sequences differ slightly by whether the actress takes a gun or not.
According to Figure \ref{fig_gundtwmatch}, DTW fails to capture this
difference, and it returns nearly ``perfect'' alignment for two semantically
different sequences.
%It cannot distinguish the pointing action and putting down action either.
Besides, it is shown that DTW un-intuitively maps one point to many points
on the other sequence.

%Some attempts have been conducted to mitigate this problem, such as
%shapeDTW, ..., ect. Discuss each solution. Despite the encouraging
%performance improvement using these solutions, there is still a warrant of
%learning representations for each time step.
Time points of a sequence have dependencies over time axis, which provides
latent regime characterizing the sequence behavior. DTW is unable to take
into consideration this information. To mitigate the problem with DTW and
motivated by the advance of representation learning, in this paper, we
propose a novel sequence alignment algorithm, dynamic state warping (DSW).
DSW efficiently converts time points on a sequence into the corresponding
hidden states, which has integrated characteristics of the past history and
the current point. In this way, the state evolving sequence may be ``aware''
of the sequential order information and encodes the generating mechanism of
original sequences. Dynamic programming technique is employed to align the
state sequences instead of the temporal points. Therefore, DSW is prone to
match time points with similar states together.

The time complexity of the state converting process for two given sequences of length $L_{Q}$ and $%
L_{C}$ is $O(L_{Q}+L_{C})$. The alignment process takes
time complexity $O(L_{Q}L_{C})$. Then the overall cost is $O(L_{Q}L_{C})$,
at the same level as that of DTW.

Our method has several advantages:
%being able to take advantage of similarity constraints;
(1) DSW explicitly takes advantages of the autocorrelation structure
characteristics of sequential data. It provides versatile and flexible
discriminative state representations for sequences;
(2) The alignment results of DSW is semantically more interpretable than plain DTW;
%compared with classical DTW approach, our method incorporates the generating mechanism information;
%(3) DSW is a general framework to take advantage of discriminative learning techniques and probabilistic models for learning state points as representations;
(3) Using one nearest neighbor classifier, DSW exhibits lower error rates
than DTW on most datasets; (4) The flexibility of DSW allows users to fine
tune the representations of sequences to capture the discriminative features
for specific tasks; (5) DSW is able to deal with uni-variate and multivariate
sequences without adjusting the algorithm. %%in the same framework.
(6) DSW shows more robustness to noise and more suitable for long sequences than
compared methods; (7) After obtaining the state sequences, the problem is
again DTW, thus advanced techniques \cite%
{rakthanmanon2012searching,jeong2011weighted} for improving the
effectiveness and efficiency of DTW is also compatible with DSW.

As a comparison, Figure \ref{fig_gundswmatch} demonstrates the alignment
result of DSW on GunPoint dataset. The parameters of DSW are determined
randomly. First of all, it is noticeable that DSW is able to match two sequences
correctly and consistently with the three true actions. Second, whether the
actress takes a gun out of the holster or holding a gun in hand, is
distinguished correctly.

Our contributions in this work include: (1) We propose a novel state
alignment algorithm for sequences. This algorithm is as efficient as DTW but
provides more interpretable and accurate alignments. (2) By combining with
one nearest neighbor classifier, our method achieves significantly better
classification results than DTW. (3) We perform extensive experiments
comparing our method with Euclidean distance (ED) and DTW. We also
empirically analyze the relating properties of the state converting
component of DSW extensively. (4) Our strategy provides a general learning
framework and it is possible to incorporate discriminative learning and
representation learning techniques.

The rest of this paper is organized as follows: in Section 2, we introduce
preliminary knowledge about time series classification, DTW and recurrent neural
networks; in Section 3, we introduce DSW in detail; Section 4 performs
extensive experiments to evaluate DSW; finally, Section 5 concludes this
paper.

\section{Background and Related Work}

In this section, we first clarify some notations and basic knowledge about
sequence learning. Then we introduce the DTW algorithm in detail. Since our
method is based on recurrent neural networks, we introduce recurrent neural
networks at last.

\subsection{Sequence Distance}

\begin{figure}[!t]
\centering
\includegraphics[width=3in]{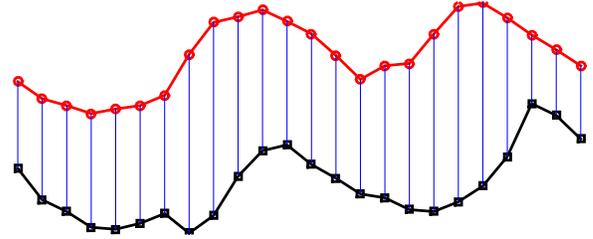}
\caption{Illustration of Euclidean distance.}
\label{fig_EDmatch}
\end{figure}

A sequence is a series of observations for at least one variable. We use a
matrix $X = [x^{1},x^{2},\cdots,x^{L_{X}}]^{T}\in R^{L_{X}\times d}$ to denote a sequence,
where $x^{i} \in R^{d\times 1} (i\in [1,L_{X}])$ is an observation at a time point indexed by $i$,
$L_{X}$ is the length of series $X$ and $d$ is the
number of variables. Each sequence is associated with
a label $y$. The task of classification is to learn a function mapping from $%
X$ to $y$.

Given two sequences $Q$ and $C$, if they are of equal length, one can easily
compute their distance using Euclidean distance (ED) or p-norm distance \cite%
{faloutsos1994fast}, such as $ED(Q,C)=\sqrt{\sum_{i=1}^{L}%
\sum_{j=1}^{d}((q^{i,j}-c^{i,j})^{2})}$. This kind of distance is also
called lock-step distance since it matches elements of sequences according
to their position (See Figure \ref{fig_EDmatch}).

However, in most real-world applications, sequences may be of varying
length. In this case, elastic distance, such as DTW and longest common
subsequence \cite{ding2008querying} etc., is used to compute the distance.

\subsubsection{Dynamic Time Warping}

\begin{figure}[!t]
\centering
\includegraphics[width=3in]{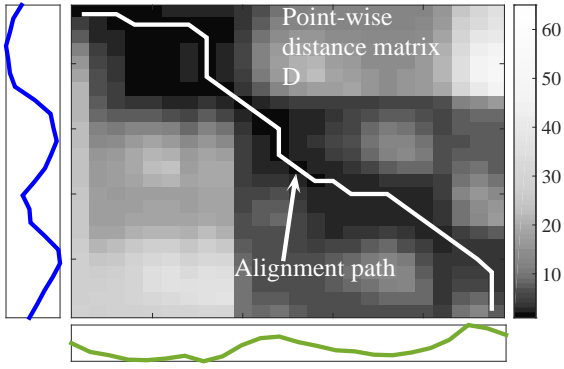}
\caption{Illustration of searching for the optimal alignment using DTW. The
two sequences are demonstrated on the left and bottom. The point-wise
distance matrix and the optimal alignment path are presented. }
\label{fig_DTWalign}
\end{figure}

Given two sequences $Q$ and $C$ of possibly different lengths, DTW stretches
or condenses two sequences to the same length. DTW allows nonlinear local
warping to take into account possible distortions. It finds an optimal alignment between $Q$ and $C$ such that the
accumulative Euclidean distance is minimized \cite{keogh2001derivative}. In
doing so, it returns a distance value that measures the similarity between
the two sequences.

Denote $D \in R^{L_{Q}\times L_{C}}$ as the distance matrix of time points
of $Q$ and $C$. DTW algorithm travels from the position $[1,1]$ to position $%
[L_{Q},L_{C}]$ of $D$ to find a sequence of indices $A$ and $B$ of the same
length $l\geq \max\{L_{Q},L_{C}\}$ such that the cumulative distance $%
\sum_{i=1}^{l}D_{A_{i},B_{i}}$ is minimized. In this way, the point $A(i)$
of sequence $Q$ is matched with the point $B(i)$ of sequence $C$. See Figure %
\ref{fig_DTWalign} for an illustration.

To be a valid alignment, the paths $A$ and $B$ have to satisfy three
constraints:

\begin{enumerate}
\item $A(1)=1, B(1)=1$

\item $A(l)=L_{Q}, B(l)=L_{C}$

\item $\forall i\in \lbrack
1,l-1],(A(i+1),B(i+1))-(A(i),B(i))=\{(0,1),(1,0),(1,1)\}$
\end{enumerate}

Hence, the alignment of two sequences should start from the first time point
and end with the last point. Each time point has at least one matching point
on the other sequence. The match increases monotonically.

Formally, denote $\alpha \in \{0,1\}^{l\times L_{Q}}$ and $\beta \in
\{0,1\}^{l\times L_{C}}$ be two warping matrices corresponding to the
warping path $A$ and $B$. Let $\alpha (i,A(i))=1$, $\beta (i,B(i))=1$, where
$i\in \{1,2,\cdots, l \} $. All other entries are zero. In this case the two
warped sequences become $\alpha \times Q$ and $\beta \times C$. The overall
cost function of DTW is generalized as:%
\begin{eqnarray}
Cost&=&\arg \min_{l,A,B }(\sum_{i=1}^{l}D_{A_{i},B_{i}})  \notag \\
&=& \arg \min_{l,\alpha ,\beta }(\alpha \times Q-\beta \times C)^{2}
\label{equ_DTW}
\end{eqnarray}
Therefore, DTW is essentially Euclidean distance except that local
distortion is allowed. The above cost function can be solved with time
complexity $O(L_{Q}L_{C})$ using dynamic programming:
\begin{equation*}  \label{equ_dynamicprogramming}
Cost(m,n)=D(m,n)+\min \left\{
\begin{array}{c}
D(m-1,n), \\
D(m-1,n-1), \\
D(m,n-1)%
\end{array}%
\right\}
\end{equation*}

\subsubsection{Related Work}

Numerous trials have been made to enhance DTW, which can be roughly divided
into two categories: adjusting the point-wise distance (e.g. \cite%
{garreau2014metric,zhou2016generalized,keogh2001derivative}) and
heuristically constraining the DTW (e.g. \cite%
{jeong2011weighted,batista2014cid}).

Garreau et al. \cite{garreau2014metric} proposed to learn a distance metric
for measuring the similarity between two points. However, this method takes
the difference between the empirical alignment and true alignment as the
cost function. Therefore, it requires the true alignments among sequences,
which is not available for many real-world applications. Besides, it is only
feasible for multi-variate sequence alignment. Zhou et al. \cite%
{zhou2016generalized} combined DTW with canonical correlation analysis
(CCA), termed canonical time warping (CTA). The point-wise distance is
linearly transformed by CCA. CTA alternatively optimizes the CCA and DTW
alignment to minimize the accumulative distance. However, CTA is only
applicable for multivariate sequences. The objective function of CTW is
non-convex and may return local minima.

%If two sequences are not apparently correlated, CTW may have the risk of over-fitting.
In \cite{petitjean2014dynamic}, Petitjean et al. proposed to replace
multiple sequences with an average sequence $\hat{T}=\arg \min_{\hat{T}%
}\{DTW(T_{i},\hat{T}) \}_{i=1}^{N}$ using the DTW distance. But it does not
have a contribution on improving the alignment of DTW algorithm. Keogh and
Pazzani \cite{keogh2001derivative} proposed to replace the coordinator
distances by the derivative distance (DDTW). Therefore, points with similar
changing trend are matched. Jeong et al. \cite{jeong2011weighted} proposed
to weight the match of two points by the length of stretch such that the
long shift is penalized (WDTW). A combination of WDTW and DDTW is also considered (WDDTW). Batista et al. \cite{batista2014cid}
proposed a similar penalty-based distance. That method additionally
considers a coefficient on top of the DTW distance in order to avoid small
distance between sequences of different complexity (CID-DTW). Like DTW,
DDTW, WDTW and CID-DTW still have the limitation in considering the
sequential nature of data. Other studies focus on improving the efficiency
of DTW \cite{rakthanmanon2012searching} but do not improve the alignment
results of DTW.

%Previous work either adjusts the matching path to penalize severe warping to
%make DTW more effective or design heuristic strategies to constrain the
%search range to make DTW efficient.
%However, little effort has been
%conducted to improve the basic subroutine of the DTW.
Our method DSW is similar to methods adjusting the point-wise distance. But
DSW is different from previous methods in that DSW is able to utilize the
sequential dynamic information of sequences. From Equation (\ref{equ_DTW}), it
is clear that given the distance matrix $D$, the basic subroutine of DTW is
the point-wise Euclidean distance of temporal axis coordinate values. The
point-to-point distance ignores local autocorrelation structure information.
This is the key bottle-neck of the classification performance of DTW. In
this paper, we are going to offer a remedy by taking advantage of the memory
ability of recurrent networks to learn a state representation for each time
point. The alignment is performed in the state space of sequences instead of
the time domain.

\subsection{Recurrent Neural Network and Reservoir Computing}

\begin{figure*}[tbp]
\centering
\includegraphics[width=6in]{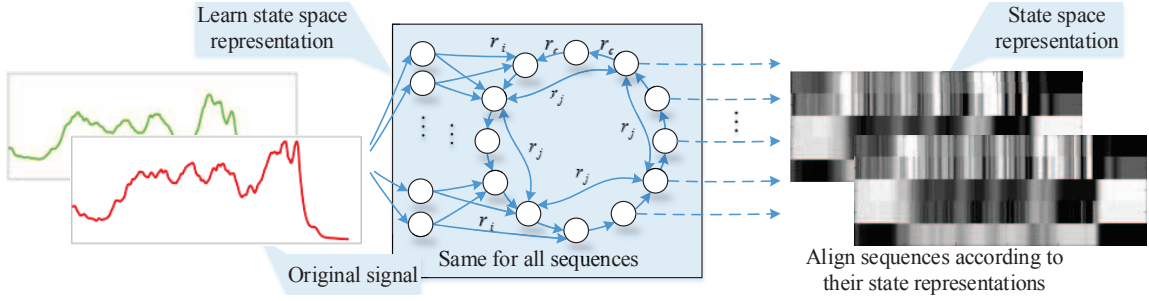}
\caption{The semantic demonstration of the key idea of DSW. The CRJ
reservoir network converts the original sequences to state trajectory
sequences. Then two points are paired if their states are similar.}
\label{CRJ}
\end{figure*}

Recurrent neural network (RNN) enables recurrent connections among neurons.
RNN learns a representation for each time point that takes into account the
previous input history and the current input. Unlike HMM, which ignored
earlier points beyond a threshold, RNNs do not constrain the memory length
\cite{haykin2009neural}. In RNN, the far earlier inputs give less influence
to the representation of the current input. This implicit fading memory has
enabled RNN a powerful tool in language modeling \cite%
{mikolov2013distributed}, sequence process \cite{pei2016modeling} and video
prediction \cite{goroshin2015unsupervised} etc.

Reservoir network is a type of RNN. It has a high dimensional recurrent
network that is fixed during training called reservoir. The reservoir
provides general dynamical features for input sequences providing it is
designed properly. In \cite{chen2014learning}, the output weights of the
reservoir model is proved to be able to extract discriminative features from
the whole sequence.
%These features are further used to learn a model for the sequence.
%Classification is performed in the model-spanned space.
It assumes that the generating mechanism of the sequences is time-invariant.
Similarly, Fisher kernel \cite{maaten2011learning} learns vectorial fisher
score as a representations for the whole sequence using a probabilistic
generative model. Fisher kernel assumes that similar objects stretch the
generative model parameters to a similar extent. In Fisher kernel, all the
sequences have to be approximated by only one generative model, which seems
to be restrictive.

This current work is different from previous studies. Our goal is to uncover
the state transition track by learning state space representations for time
points, rather than a representation of the whole sequence. Thus the
assumption of the time-invariant generating mechanism is not a limitation
for DSW. Besides, it is critical for model space learning method to learn
faithful models for original sequences. So a large reservoir is usually
needed, which adds computational burdens. For more information about RNN and
reservoir, we refer the readers to \cite{lukovsevivcius2009reservoir}.

DTW is commonly used comparing time points in the time domain in previous
sequence classification methods, which makes it sensitive to noise and
weakly interpretable \cite{ye2009time}. Numerous methods have been proposed
to improve the effectiveness or efficiency of DTW.
%Other state-of-the-art methods perform feature extraction methods for a total sequence. These works either handcraft the features \cite{baydogan2013bag} or use a probabilistic generative model to learn the features \cite{maaten2011learning}. They are limited in taking advantage of the generating mechanism of sequences or the available similarity structure constraints.
However, little prior work has contributed from the aspect of learning
point-wise representations to enhance DTW for sequence classification.
%Other state-of-the-art methods perform feature extraction methods for a total sequence. These works either handcraft the features \cite{keogh2001derivative} or use a probabilistic generative model to learn the features \cite{maaten2011learning}. They are limited in taking advantage of the generating mechanism of sequences or the available similarity structure constraints. Little prior work has contributed from the aspect of learning point-wise representations to enhance DTW for sequence classification.

As a remedy, in this work, we propose to learn temporal point
representations for sequences. We learn flexible and versatile
representations for each time point. The integrated information in the time
point representations provides discriminative features. The proposed method
makes use of the history information of sequences to make the alignment
semantically more sensible and provides better classification accuracy in
comparison with DTW.

%Bag of features.
%Reservoir kernel. Existing methods need expert knowledge to design
%domain-specific local features, while our method is completely automatic.

%Out work is different from previous studies. Previous studies only describe
%local features as a descriptor. However,

\section{Dynamic State Warping}

This section introduces the dynamic state warping method.
Sequences are structured objects that have correlation among different time
points. DTW finds an optimal alignment between two sequences so that the
cumulative Euclidean distance is minimized. However, it has limitations in
making use of the dynamic information.

We propose an algorithm DSW as a mitigation. In particular, the mechanism of
DSW is a two-stage process. In the first stage, DSW uses the reservoir
network as a general purpose temporal filter to convert original sequences
into state sequences, which encapsulates the generating mechanism of original
sequences. In the second stage, the same alignment operation as that of DTW
is performed to align the state trajectory sequences.

The motivations of choosing the reservoir network as the
signal-to-state converting model include: (1) it provides parsimonious state
representations for sequences points efficiently; (2) it is able to take
advantage of the sequential autocorrelation information of sequences; (3) it
requires much less effort to tune the network compared with a classical
neural network.

Echo state network (ESN) is a kind of reservoir model. ESN is characterized
by a non-trainable high dimensional nonlinear dynamical reservoir and an
efficiently trained linear readout layer. The reservoir is randomly
generated under the constraint of the maximum eigenvalue being less than
one. This is also called echo state property. Loosely speaking, it requires
that the initial inputs have little influence on the final state. It is
usually implemented by first normalizing the reservoir weight matrix to have
unitary spectral radius. Then we multiply it with a scaling parameter. The
readout layer assembles the state space features of the reservoir to learn a
function mapping from reservoir state to the output sequence. Linear
regression is usually employed to learn the functions.

In this study, we are not going to count on the assembling ability of the
readout layer, but we will analyze its effect on the performance of DSW in
the experiment (See Section \ref{subsub_predicablity}). The approximation ability of the
ESN reservoir helps leverage the autocorrelation of original sequences.

To justify the randomness of ESN, Rodan et al. \cite{rodan2012simple}
propose a topologically fixed reservoir, cycle reservoir with jumps (CRJ).
CRJ reservoir is more constrained than ESN reservoir. It connects the
reservoir neurons in a uni-directional circle and allows fixed length jumps
on the circle. In this case, there are only three kind of connections for
CRJ reservoir, i.e. input, jump and cyclic connections. The network weights
can be determined by three values $r=\{r_{i},r_{j},r_{c} \}$ for each kind
of connection.

This paper takes CRJ as the base model for ease of analysis. The DSW
algorithm is demonstrated in Algorithm \ref{algorithm_DSW}. Figure \ref{CRJ}
gives a sematic illustration of the key idea of this paper.

\begin{algorithm}
\caption{Dynamic state warping}
\label{algorithm_DSW}
\begin{algorithmic}[1]
\STATE \textbf{Input:} Two sequences $Q\in R^{L_{Q}\times d},C\in R^{L_{C}\times d}$; \#neurons of the reservoir; jump length of reservoir; initial parameters for network ($\{r_{i},r_{c},r_{j}\}$).
\STATE \textbf{Output:} A distance between the two sequences.
\STATE Use the CRJ network to convert the original sequences into reservoir state space.
\STATE Compute state point level distance matrix $D$.
\STATE Search the optimal warped path $\alpha$ and $\beta$ using dynamic programming.
\STATE Return the accumulative distance (Equation (\ref{equ_DTW})) as the distance between two input sequences.
\end{algorithmic}
\end{algorithm}

The form of CRJ is generalized as:
\begin{equation}  \label{equ_reservoir}
\begin{cases}
\mathbf{S}(t+1)=g(\mathbf{R}S(t)+\mathbf{V}X(t+1)) \\
f(t)=g_{out}(\mathbf{W}S(t)+b)%
\end{cases}%
\end{equation}
where $S(t) \in R^{N}$ is the reservoir state, $N$ is the number of
neurons in the reservoir; $X(t) \in R^{n}$ is the input sequence,
%with an additional bias term,
$n$ is the number of input neurons; $\mathbf{R}%
\in R^{N\times N}$ is the reservoir weight matrix, $\mathbf{V}\in R^{N\times
n}$ is the input weight matrix, $\mathbf{W}\in R^{O\times N}$ is the output
weight matrix, $b \in R^{O}$ is the bias term, $O$ is the number of output
neurons; $g$ is the state transition function, which is a nonlinear function
and usually taken as $tanh$ or the sigmoid function. $g_{out}$ is the
activation function of outputs. Without loss of generality, in this paper, we fix $g=tanh$ and $g_{out}$
as identity function.

Given an input sequence $X\in R^{L_{X}\times d}$, we first drive it through
the nonlinear dynamic reservoir and obtain a state transition trajectory
sequence $S\in [-1,1]^{L_{X}\times N}$, where $N$ is the size of reservoir.
It transforms from the previous state to a new state in the state space
given a new input time point. In this way, the temporal signal space
sequence is converted into multi-variate state space sequences. The state
trajectory sequence encapsulates the generating mechanism of the original
sequence by taking into consideration the previous state and current input
(Equation (\ref{equ_reservoir})). In particular, the reservoir state
representation is the activation of reservoir neurons with different driving
input sequences. The state space representations provide discriminative
features with more versatility and flexibility than original sequences.

We then employ DTW to align the state trajectory sequences as usual. It prefers
to align points on two sequences with similar states.
%The DSW is summarized in Algorithm \ref{algorithm_DSW}.

\begin{figure*}[!t]
\centering
\includegraphics[width=6.5in]{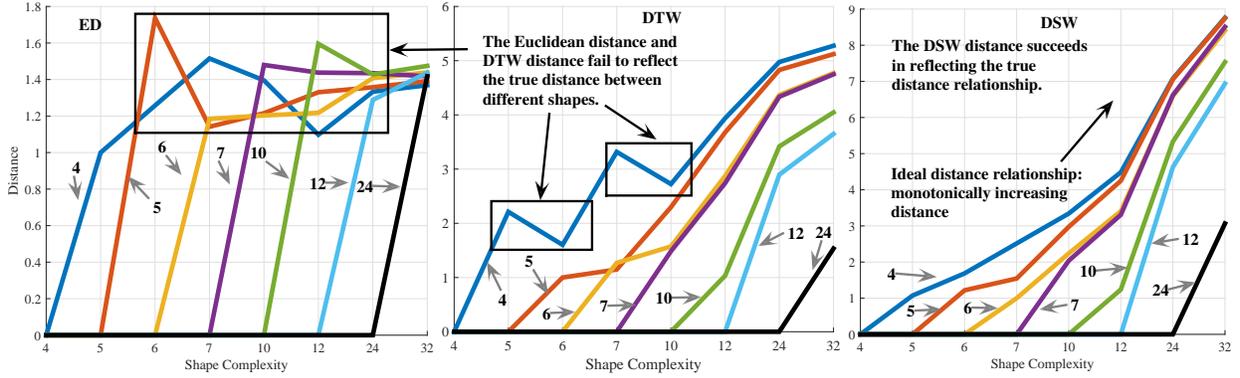}
\caption{Distance among shapes. The ylabel is the distance between the
shapes. The xlabel indicates the shape labels. Each line indicates distance
values of a shape to other shapes. Shapes with similar endpoints should
share a small distance, while significantly different shapes share a large
distance. It shows DSW approximate the true distance relationship well.}
\label{fig_shapedistance}
\end{figure*}

Therefore, the difference between DTW and DSW lies in that DTW uses distance
between time points as a subroutine, while DSW uses the distance of the state
space representations.

Since our concern in this paper is to learn point-level states as
representations, our framework can be naturally generalized to time-variant
and time-invariant, uni-variate and multi-variate series.
%Our work is a general purpose framework that one can include  as a subset.
%DSW takes advantage of the memory ability of neural network in providing more
%versatile discriminative features for each time point.

\textbf{Proposition:} The state sequence is able to scale the noise in the original sequence by a constant scaling.

Proof: The state sequence is able to reduce the noise in the original signal by constraining $r_{i}^{2}$.
Given an additive noise $\varepsilon $ in the sequence $X$, the distance between
the state sequences is then:%
\begin{eqnarray*}
&&||S^{\varepsilon }(t)-S(t)||^{2} \\
&=&||g(\mathbf{R}S(t)+\mathbf{V}%
(X(t)+\varepsilon ))-g(\mathbf{R}S(t)+\mathbf{V}X(t))||^{2} \\
&\leq &||\mathbf{R}S(t)+\mathbf{V}(X(t)+\varepsilon )-\mathbf{R}S(t)-\mathbf{%
V}X(t)||^{2} \\
&=&||\mathbf{V}\varepsilon ||^{2}\leq r_{i}^{2}||\varepsilon ||^{2}%\leq ||\varepsilon ||^{2}
\end{eqnarray*}
where $S^{\varepsilon }$ is the noisy state sequence.
The above equation reveals that the noise in the state sequence is scaled by the input weight $r_{i}^{2}$.
The nonlinear state transition function $g$ can be $tanh$ or $sigmoid$ function.
Note that their derivative is not larger than one, thus we have $tanh(\delta)\leq \delta$.
That is, the noise level is reduced in the state sequence by adjusting $r_{i}$ (usually $r_{i}\leq 1$ in the typical ESN setting \cite{lukovsevivcius2009reservoir}).

\section{Experiment}

\label{section_Experiment}

In this section, we perform extensive experiments to evaluate our method.
Since all the recent enhancement for DTW also works in DSW, our strategy is to compare DSW with DTW and ED primarily.
We also compare DSW with more advanced algorithms for classification performance.
In particular, this section is divided into four parts. We first evaluate our
method on synthetic datasets to show its effectiveness as a distance metric
(subsection \ref{sec_distancemeasure}). Secondly, the robustness is evaluated on
synthetic noisy data. The scalability of long series is tested by
consistently increasing the length of sequences (subsection \ref%
{sec_robustscalability}). Thirdly, we evaluate DSW by the classification
results on 85 UCR time series datasets \cite{UCRArchive} in comparison with
ED, DTW and other algorithms (subsection \ref{sec_stateoftheart} and \ref%
{sec_texas}). Finally, we analyze the relating properties of our method to
provide more insights into DSW (subsection \ref{sec_DSWproperty}).

\subsection{Experimental Setup}

We compare DSW with Euclidean distance and DTW.
%To see whether the combination of state representation and original sequence can help improve the performance, we also include this case as a comparison, denoted by DSWin.
For dynamic programming process, the size of warping window is not
optimized. However, since the second stage of our method is intrinsically
DTW, more advanced strengthening techniques for DTW, such as those proposed
in \cite{rakthanmanon2012searching} and \cite{jeong2011weighted}, can be
easily incorporated. Without explicit mention, the reservoir of DSW is
randomly generated. The sequences have been normalized to have zero mean and
unit standard deviation. %The data, detailed numerical results and
%The source code of DSW is available from our website\footnote{%
%www.github.com}.

\subsection{Distance Measurement}

\label{sec_distancemeasure}

\begin{figure}[!t]
\centering
\includegraphics[width=3.3in]{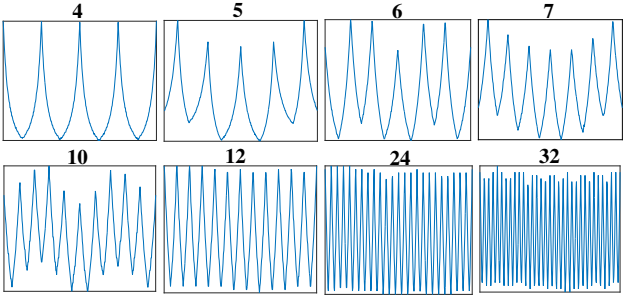}
\caption{Illustration of sequences of shapes with different number of
endpoints.}
\label{fig_shapesequence}
\end{figure}

\begin{figure}[!t]
\centering
\includegraphics[width=3.3in]{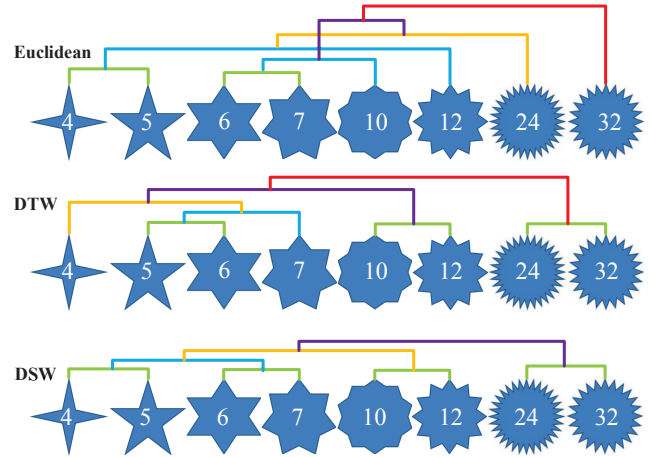}
\caption{Clustering results of shape sequences using Euclidean distance, DTW
and DSW. DSW clusters shape sequences in a manner consistent with human's
intuition. This result indicates that DSW is indeed effective in capturing
the true similarity structure among sequences.}
\label{fig_cluster}
\end{figure}

To demonstrate that DSW is a distance measurement that can approximate the
true similarity structure of sequences, we compare DSW with Euclidean
distance and DTW on a synthetic data. In particular, we consider the shapes
of polygons \cite{batista2014cid}, see Figure \ref{fig_cluster} for
an illustration. The number of endpoints of a shape is taken as the label.
For each polygon the original shape in two dimensional space is converted
into one dimension series by computing the distance from the center to edge
points. The resulting series is demonstrated in Figure \ref%
{fig_shapesequence}.

%Table \ref{table_ED}, \ref{table_DTW} and \ref{table_DSW} shows
Figure \ref{fig_shapedistance} demonstrates the distance among different
shapes using Euclidean distance, DTW and DSW respectively. According to Figure \ref{fig_shapedistance}, compared with
ED and DTW, DSW captures the similarity between different shapes.
Ideally, the distance between two shapes should increase with the difference of their
complexity. Euclidean distance cannot approximate the true similarity
relationship between polygons. DTW is better than Euclidean distance but
fails in capturing the similarity between shape 4 and other shapes.

We also show the hierarchical clustering results of three distance
measurements. As illustrated in Figure \ref{fig_cluster}, DSW generates
clustering result that is more consistent with the semantic meaning.

%\subsubsection{Scaling}
%
%\subsubsection{Stretching}

\subsection{Robustness and Scalability}

\label{sec_robustscalability}

\subsubsection{Robustness}

\begin{figure}[!t]
\centering
\includegraphics[width=3.3in]{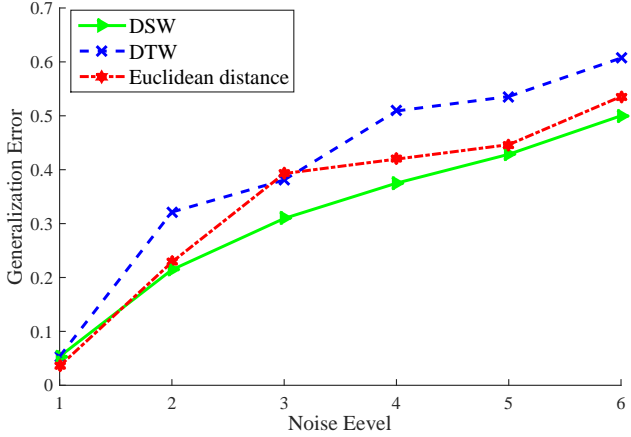}
\caption{The Generalization performance of Euclidean, DTW and DSW distance
with 1NN classifier when facing noisy data. This result demonstrates that
DSW is more robust than Euclidean distance and DTW.}
\label{fig_noiserobust}
\end{figure}

%To test the robustness of different distance measurements, we increasingly
%add noise on the generated synthetic dataset using a gaussian
%distribution with zero mean and standard deviation ranging in $[0.1,0.5]$.
%The experiment is repeated 5 times. In each run, the synthetic data is
%re-generated then noise is added. The average result is demonstrated in Figure \ref%
%{fig_noiserobust}.

To evaluate the robustness of different distance measurements, we conduct
experiments varying the noise in the dataset. In detail, we choose a
benchmark dataset Coffee \cite{UCRArchive}. On this dataset, the DTW and Euclidean distance
achieve zero error rate when no additional noise is incorporated.
We add a Gaussian white noise into coffee dataset with zero mean and standard deviation
varying in the set \{ 0.1,0.3,0.5,0.7,0.9,1.1\}. The standard deviation is
set in this manner in order to clearly present the tendency. On each noise
level, we perform 10 runs and report the average. The generalization error
rate is computed using 1NN classifier with different distance measurement.

The results are presented in Figure \ref{fig_noiserobust}. It shows that DSW
is more robust to noise than competitive methods. DSW consistently achieves
smaller generalization error rates than DTW, except that they are equal when
noise level is 0.1. DTW is sensitive to noise because its basic unit is to
compute point-wise Euclidean distance between points and counts heavily on
the sequence shape during alignment. Yet, the Gaussian noise changes the
magnitude of original sequences. This partially explains the noise
sensitivity of DTW. Euclidean distance maintains intermediate performance.
We also observe that when the noise level is low, Euclidean distance
performs a little better than DTW and DSW distance.

\subsubsection{Scalability to Long Sequences}

\begin{figure}[!t]
\centering
\includegraphics[width=3.3in]{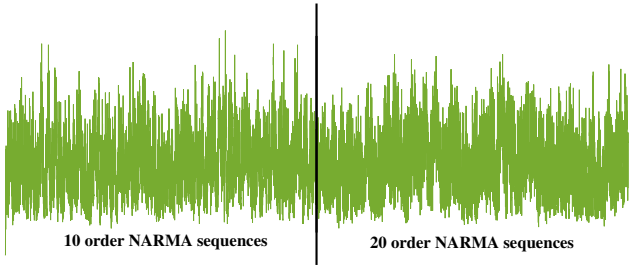}
\caption{Illustration of 10 order and 20 order NARMA synthetic sequences.}
\label{fig_NARMA_demosequences}
\end{figure}

\begin{figure}[!t]
\centering
\includegraphics[width=3.3in]{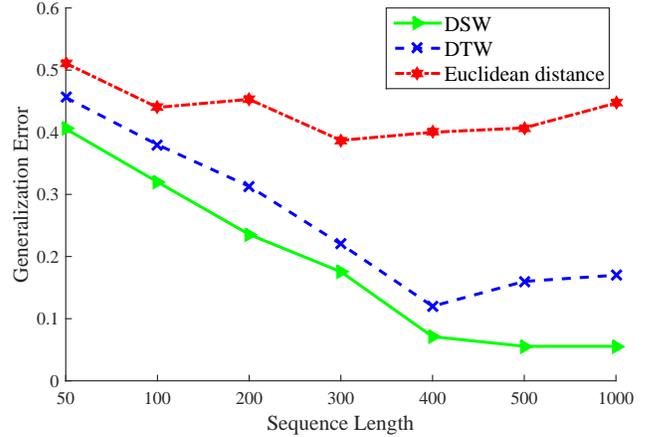}
\caption{The generalization trend of three distance measurements facing time
series of different length.}
\label{fig_sequencelength}
\end{figure}

\begin{figure*}[!t]
\centering
\includegraphics[width=6.5in]{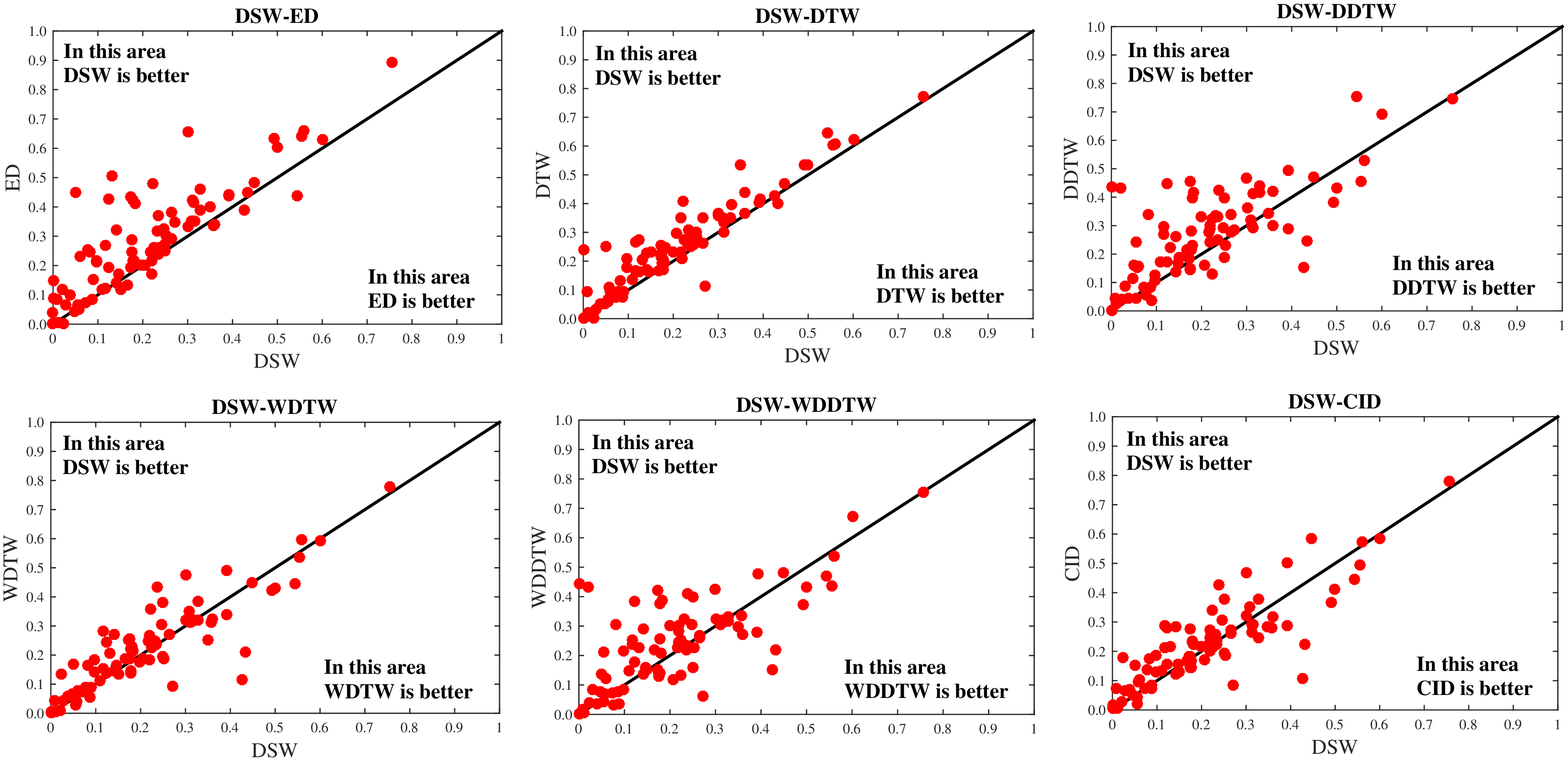}
\caption{The pairwise comparison of 1NN classification performance of DSW,
ED, DTW, DDTW, WDTW, WDDTW and CID-DTW on 85 UCR time series datasets.}
\label{fig_pairwise}
\end{figure*}

In this subsection, we employ synthetic datasets containing varying length
sequences to evaluate the scalability of DSW.

In particular, we generate a series of 10 order and 20 order NARMA
sequences:
\begin{eqnarray*}
s(t+1) &=&0.3s(t)+0.05s(t)\sum_{i=0}^{9}s(t-i)+ \\
&&1.5u(t-9)u(t)+0.1 \\
s(t+1) &=& tanh(0.3y(t)+0.05y(t)\sum_{i=0}^{19}y(t-i)+ \\
&& 1.5u(t-19)u(t) + 0.01) + 0.2
\end{eqnarray*}
where $s(t)$ is the output sequence, $u(t)$ is the input sequence, $u(t)$ is
independently and identically generated in the range [0,0.5] according to
uniform distribution. These two kinds of sequences are generated using the
same input sequence. The 10 order and 20 order NARMA sequences are treated
as two classes. The original sequences are illustrated in Figure \ref%
{fig_NARMA_demosequences}.

We use the NARMA model of 10 order and 20 order respectively, to generate
synthetic data with varying sequence length. This experiment aims to present
what will happen when sequence length is becoming longer and longer. In
detail, we generate 10 order and 20 order NARMA sequences of the length in
\{5000,10000,20000,30000,40000,50000,100000\}. For each length, the
generated sequences are then divided into non-overlapping subsequences to
obtain 100 subsequences. We randomly select 25 sequences of 10 order and 25
sequences of 20 order as the training set of size 50. The rest 50
subsequences are used as test set. We use 1NN classifier to classify the
test set using the training set. For each length, the synthetic dataset is
generated 10 times and the classification is also repeated 10 times. The
average generalization error rate is collected as the final result.

Figure \ref{fig_sequencelength} demonstrates the performance of different
distance measurements in terms of varying sequence length. From Figure \ref%
{fig_sequencelength}, we can make three main observations:
(1) It shows that Euclidean distance scales poorly to long sequences. When the sequence length
exceeds $400$, Euclidean distance begins to degrade the generalization
performance of 1NN classifier. This result is as expected and consistent
with our intuition that Euclidean distance is not suitable for
long series.
(2) The performance of DTW initially improves when facing
long sequences. Its performance is much better than Euclidean distance. This
result explains why DTW is so successful in sequence processing domains \cite%
{rakthanmanon2012searching,bagnall2016great}. Sequences are usually long
and the time points are usually much larger than the number of observations. However, when
sequence length grows longer than 600 points, the performance of DTW begins to
worsen. (3) Unlike the competitive distance measurements, DSW is
consistently improving the performance in our experiment.
%In particular, DSW performs a little better than DSWin when
%sequence length is short. With longer sequences, DSWin does a little better
%than DSW.
%This interesting observation is consistent the classification
%result on UCR benchmark datasets.
%In this manner, we recommend to use DSW
%for short sequences classification tasks, and DSWin for long sequence
%classification tasks.

%\subsection{Basic Error Rate and Time Complexity}

\subsection{Classification Performance on Benchmark Datasets}

\label{sec_stateoftheart}

\begin{table}[tbp]
\caption{Number of datasets on which DSW perform better than compared methods using 1NN classifier.}
\label{Table_pairwisewins}\centering
\begin{tabular}{c|cccccc}
\hline
    & ED & DTW & DDTW & WDTW & WDDTW & CID \\ \hline
DSW & 70 & 74  &  62  &  55  &  51   &  51
\\ \hline
\end{tabular}%
\par
\begin{flushleft}
The numerical value means the number of wins of DSW in
comparison with a compared method on 85 datasets. It is observed that
DSW is better than compared methods on most datasets. Note that the derivative, weight and complexity penalty are easy to
be incorporated in DSW, which may further enhance the performance of DSW.
\end{flushleft}
\end{table}

\textbf{Datasets} Our experiments are performed using the UCR time series
datasets \cite{UCRArchive}. The 85 UCR datasets \cite{UCRArchive} are
collected from different domains such as insect recognition, medicine,
engineering, motion tracking, image and synthetic data etc. The datasets
have already been divided into training and test set. The length of
sequences varies from one dataset to another, with the minimum length 24 and
maximum length 2709. In each dataset, the sequences are of equal length. The
size of datasets varies from 24 to 8926. The number of classes varies from 2
to 60. Detailed information about the datasets is available on the website
\cite{UCRArchive}.

\textbf{Experiment Setup and Parameter settings} Our results are averages
over 10 repetitions. In each repetition, the network is optimized by
selecting one network from 20 randomly initialized networks using
leave-one-out cross validation (LOOCV). The input dimensionality of
reservoir network is set as 2. The reservoir size is set as 5. The spectral
radius scaling is set as 0.85. This setup is applied to all datasets.

In our experiment, the network is selected in this arbitrary manner. Thus
the results may not be optimal for our method. We can of course design
specific networks for different datasets by using the result of following
subsections \ref{subsection_reservoiranalysis}. We do not optimize the
classification result using more advanced strategy to simplify operations.
However, when dealing with real-world problems, one can of course choose a
better parameter setting using simple search strategies e.g. grid search.
For example, on BeetleFly dataset, we search the eigenspectral scaling using
the range reported in subsection \ref{subsub_spectral} and improve the
accuracy of DSW by 15\%. Indeed, our results show that even using randomly
generated reservoir and our LOOCV strategy can yield surprisingly good
results.

The Euclidean distance and DTW algorithm are determined thus only one run is
performed on each dataset.

\textbf{Compared methods} The seven compared method are:

\textbf{(1) ED}: Each of the test sequences is compared to the sequences
in the training set by Euclidean distance. Then the label of the nearest
neighbor of the query is returned as the prediction label.

\textbf{(2) DTW}: For each sequence in test set, we compute a distance
between this sequence and all training sequences. The distance is computed
using DTW. %\footnote{The code can be obtained from the website.}.
Then we scan all the training set to find a training sequence that is
nearest to the given query. The predicted label is taken as the label of the
nearest sequence in the training set.

\textbf{(3) DDTW}: The first order distance is calculated for DTW as in \cite{keogh2001derivative}. Points with similar derivative
are more likely to be matched.

\textbf{(4) WDTW}: A multiplicative weight penalty is incorporated to reduce the warping degree. The distance between two points
on two series is then $w_{|i-j|}(Q^{i}-C^{J})^{2}$. Following \cite{jeong2011weighted}, we employ a logistic function for the weight, i.e., $w_{a}=\frac{1}{1+g(a-L/2)}$, where $a=1,2,\cdots,L, L$ is the sequence length, $g$ is set on a validation set.

\textbf{(5) WDDTW}: A combination of DDTW and WDTW, which considers the derivative and weight penalty simultaneously \cite{jeong2011weighted}.

\textbf{(6) CID-DTW}: As indicated in \cite{batista2014cid}, a complexity penalty is multiplied on the original DTW distance to take into account the difference in the complexity of two series.

\textbf{(7) DSW}: The training sequences are converted into state
sequences using a reservoir network. When a query sequence arrives, it is
first converted into state sequence using the same network. The distance
between the query sequence and training sequences is computed by DTW using
the state sequences. The network parameters e.g. the connection weights, are
initialized randomly.

%\textbf{NNDSWin}: It is the same as the NNDSW except that the signal space sequence is considered as well when computing point-wise distance.

%The running time is the clock time of a 2 Intel Xeon Quad-Core E5620 2.40GHz
%with 12GB 1333MHz DDR3 ECC Unbuffered RAM (using just one core unless
%otherwise explicitly stated).
\textbf{Platform} The experimental environment is Matlab 2015a on an Intel
Pentium Quad-Core G2120 3.10GHz CPU with 4GB RAM.

%Table \ref{table_ucrerror} presents the classification error rates on 85
%datasets. The mean value and standard deviation (in the bracket) are
%presented for DSW. The best results are boldfaced. Performance improvement
%more than 10\% is marked by the superscript $^{\ast }$.

%Figure \ref{fig_pairwise} presents the pairwise comparison of DSW with comparative algorithms on 85 UCR datasets.
%The mean value of DSW is employed for comparison.

Using the parameter-free 1NN classifier enables us to compare the
generalization error rates of DSW with compared algorithms in a pair-wise
fashion. The classification result on 85 UCR datasets is presented in Figure \ref{fig_pairwise}.

It is clear that out method achieves lower classification error rates than ED and DTW.
DSW also performs favorably than more advanced algorithms.
In detail, DSW achieves better performance on 70 out of 85 datasets than
Euclidean distance. It has better performance on 74 datasets than DTW.
The number of datasets on which DSW performs better than compared methods are reported in Table \ref{Table_pairwisewins}
.%DSWin does better than Euclidean distance and DTW on 69 and 64 datasets.
%DSW has lower error rates than DSWin on 58 datasets.

\begin{figure}[!t]
\centering
\includegraphics[width=3in]{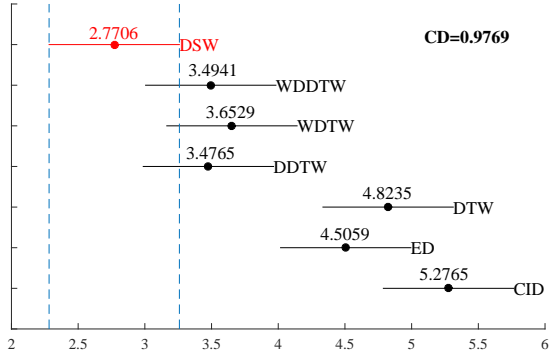}
\caption{Critical difference diagram for ED, DTW, WDTW, DDTW, WDDTW, CID-DTW and DSW.
The numerical values indicates the average rank on $85$ UCR datasets.
%ED and DTW are not significantly different %according
%to the Nemenyi test under the significance level $0.05$.
%The bold line means the connected methods are not significantly different according
%to the Nemenyi test under the significance level 0.05.
The difference is significant for algorithms with non-overlapped CDs.
DSW performs significantly better than ED and DTW according
to the Nemenyi test under the significance level $0.05$. }
\label{fig_NemenyiTest}
\end{figure}

\begin{figure*}[!t]
\centering
\includegraphics[width=6.5in]{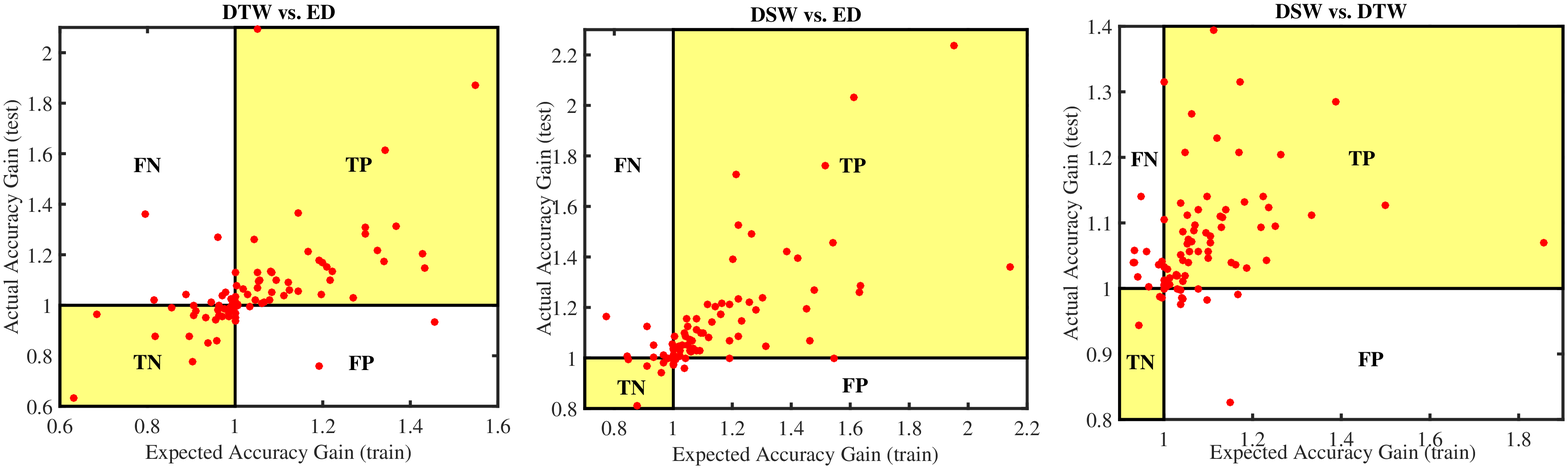}
\caption{Texas sharpshooter plot: expected performance gain on the training
set vs. actual performance gain on the test set. On most datasets, we can
correctly predict the performance of DSW. TP: performance gain on both
training and test set; TN: performance reduction on both training and test
set; FN: performance reduction on training set but performance gain on test
set; FP: performance gain on training set but performance reduction on test
set. The FP region means that DSW is overoptimistic about its performance
and is not desired. We only observe $4/85$ datasets fall into FP region for
DSW vs. Euclidean distance and $7/85$ datasets fall into FP region for DSW
vs. DTW. In FP region, most datasets are close to $[1,1]$, which indicates
the reduction is very small. }
\label{fig_shooter}
\end{figure*}

\textbf{Statistical Analysis} To provide more insights into the performance
of our method compared with ED and DTW, etc., we perform
statistical significance test for the difference of all approaches. In
detail, the generalization error rates of the compared methods on all
85 UCR datasets are first converted into ranks. Then we average the ranks across
all datasets. We employ the Friedman test to compare the different distance
measurements. The Friedman test indicates that the three distance
measurements indeed behave differently. A post-hoc pairwise Nemenyi test is
performed to evaluate the significance of the rank differences. The result
is demonstrated in Figure \ref{fig_NemenyiTest}. The main observation is
that DSW gets the lowest rank and is the best distance measurement of the
compared methods for classification. We also observe that our method is significantly better that
ED and DTW. In addition, the
difference is not significant between Euclidean distance and DTW under the significance level
0.05.

\subsection{Texas Sharpshooter Plot}

\label{sec_texas}

We have evaluated the performance of DSW compared with Euclidean distance
and DTW. The result is encouraging. Note that, it would make no sense if we
cannot know ahead of time whether DSW will perform well on a given dataset
\cite{batista2014cid}. For this purpose, we employ the Texas
sharpshooter plot to visualize if DSW is useful by predicting the
generalization ability using the classification results on the training set.

Let us take algorithm \textit{A} and algorithm \textit{B} as an example. In
detail, we use the LOOCV performance on the training set of algorithms
\textit{A} and \textit{B} to compute the expected performance gain: $\text{%
training accuracy of } \mathit{A}/ \text{training accuracy of } \mathit{B}$;
we use the accuracy on the test set to compute the actual performance gain: $%
\text{test accuracy of } \mathit{A}/ \text{test accuracy of } \mathit{B}$.
The result for Euclidean distance, DTW and DSW are presented in Figure \ref%
{fig_shooter}. The true positive (TP) region represents datasets on which we
predict a performance gain and are correct. The true negative (TN ) region
represents datasets on which we predict a performance reduction and are
correct. The false negative (FN) region represents datasets on which we
predict a performance reduction but it actually turns out to be a
performance gain. The false positive (FP) region represents datasets on
which we predict a performance gain but actually observe a performance
reduction. The FP region is not desired. From Figure \ref{fig_shooter}, we
observe that 4.7\% %and 7.1\% and DSWin vs. ED and DSWin vs. DTW and 12.9\%
points are in the FP region of DSW vs. ED; 8.2\% points are in the region of
FP for DSW vs. DTW. This result indicates that we have a very low
probability to be overoptimistic on our prediction. Besides, note that most
points in FP are close to point (1,1), which means the performance reduction
is very small.

%\subsection{Sensitivity Analysis}
%
%\subsubsection{Reservoir Size}
%
%\subsubsection{Input Dimensions}

Up till now, we have compared DSW with Euclidean distance and DTW to
demonstrate the effectiveness of our method for sequence classification.
Next we will empirically analyze the influence of the relating properties of
reservoir model on DSW.

\subsection{What is happening in the reservoir?}

\label{sec_DSWproperty} \label{subsection_reservoiranalysis}
%plot the PCA of reservoir representation distance matrix. It shows the reservoir state representation is more separated %than the original signal.
%\subsection{Are there any tricks in design the reservoir network of DSW?}
The skeptical readers may be wondering how a randomly generated reservoir
network could achieve excellent classification performance. In this
subsection, we are going to empirically uncover some insights about the
reservoir network of DSW. In particular, we will analyze the spectral radius
scaling of reservoir (\ref{subsub_spectral}), the input connection weights (%
\ref{subsub_inputweights}), the predictability of reservoir for input
sequences (\ref{subsub_predicablity}), the size of reservoir (\ref%
{subsub_reservoirsize}) and the input dimensionality (\ref{subsubInputDim}).
Our strategy to evaluate these properties is to fix the other parameters and
only vary the target property. The generalization error of DSW is calculated by the 1NN classifier.

\subsubsection{Spectral Radius Scaling}

\label{subsub_spectral}

\begin{figure*}[!t]
\centering
\includegraphics[width=6.5in]{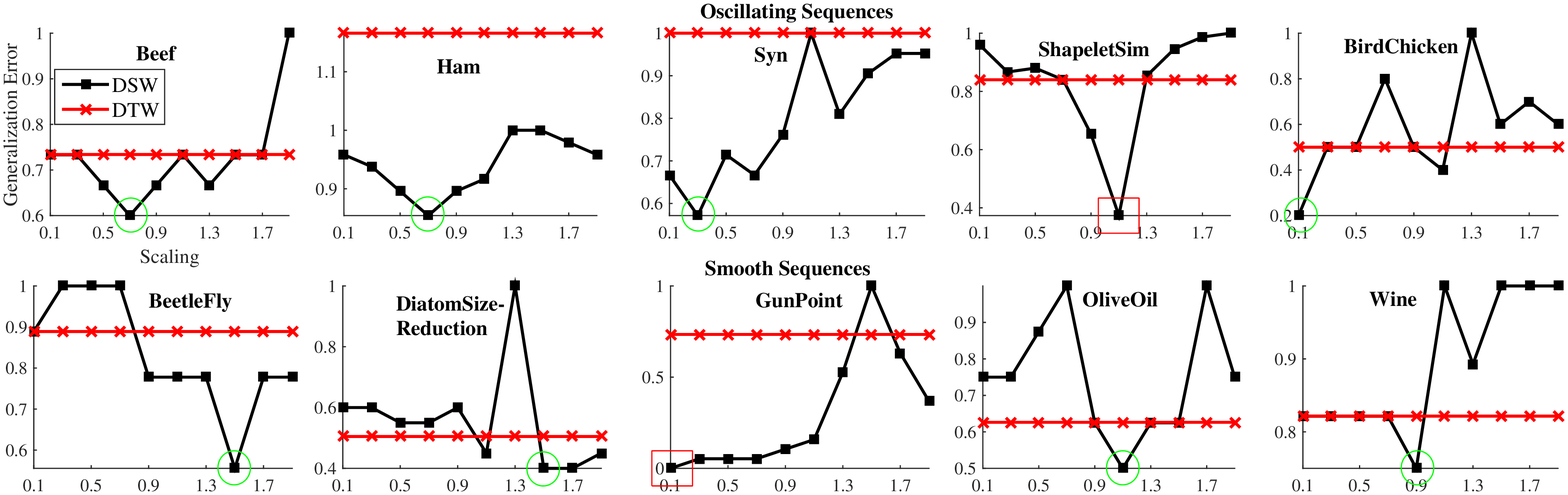}
\caption{Generalization error with varying reservoir scalings. The original error rates on each dataset
have been divided by the maxima error rate for better visualization. The top
row shows the result of fast oscillating sequences and the bottom row shows
the result of the smooth sequences. It
demonstrates that (1) oscillating sequences usually need a small scaling and smooth
sequences usually need a large scaling (see green circle); (2) local discriminative features need small scaling and global discriminative features need large scaling (see red rectangle). In addition, we plot the DTW
generalization error for comparison. Note that DSW's parameters are set
arbitrarily without optimizing by cross-validation. Despite the pessimistic setup of DSW, on most dataset DSW maintains lower error than DTW.}
\label{fig_spectralradius}
\end{figure*}

\begin{figure}[!t]
\centering
\includegraphics[width=3.3in]{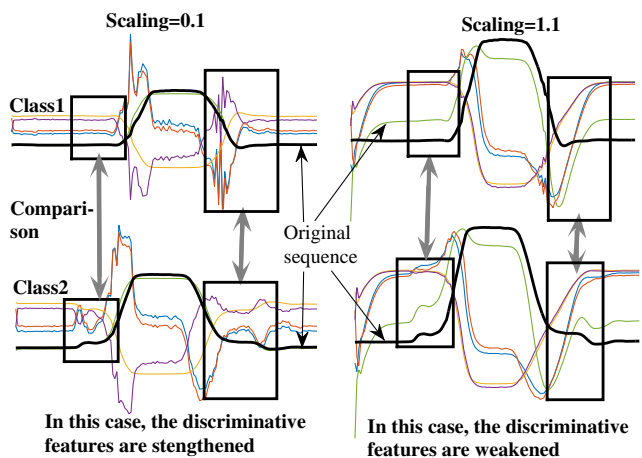}
\caption{State sequences (colored) of original sequences (black) in GunPoint dataset.
The scaling parameters are 0.1(left) and 1.1(right).
The box shows the local discriminative features of two sequences (upside and
bottom). Small scaling is good for local discriminative features. Large
scaling is good for global discriminative features. Remind that the
discriminative features for GunPoint dataset lies in a local area (see the
box). The figure demonstrates that small scaling strengthens the discriminative features (left). Yet large
scaling makes the discriminative features weaken thus hides them (right).}
\label{fig_gunexplain}
\end{figure}

Previous studies have revealed that the spectral radius scaling should be
less than one to guarantee echo state property \cite%
{lukovsevivcius2009reservoir}. That is sequence modeling should be not
sensitive to the initial value of the sequence. In particular, we have $%
\mathbf{R}=scaling\times \mathbf{R}/max(eig(\mathbf{R})) $, where $\mathbf{R}
$ is the reservoir weight matrix \cite{lukovsevivcius2009reservoir}.

To study the effect of the scaling on the classification performance of DSW,
we fix the reservoir parameters and then vary the spectral scaling
parameter. In detail, let $r_{i}=0.2$, $r_{c}=0.5$, $r_{j}=0.4$, $N=5$ and $%
jump length = 2$. The scaling parameter varies from 0.1 to 1.9 with step size
0.2.

%We examine the datasets closely and find those datasets containing severely oscillating sequences usually prefer small %scaling and more smooth datasets prefer larger scaling. Figure \ref shows the experiment results.

We artificially divide some UCR datasets into two groups according to
whether the sequences are fast oscillating ones or slowly changing ones.
Fast oscillating series changes rapidly over time, while the opposite is
slowly changing series that evolve smoothly over time. The separation of the two groups is
performed visually since it is difficult to give a definition to separate
them. The aim of doing so will be clear later.

Generally, there is a trade-off between the input and the previous state in
learning the current state representation. In particular, smaller scaling
parameter weights the previous state less and gives more importance to the
current input, resulting in short short-term memory \cite%
{lukovsevivcius2009reservoir}. On the opposite, large scaling endows more
influence to previous state, which leads to long short-term memory.
Oscillating sequences usually need a short short-term memory to model the
fast dynamics. Thus small scaling parameter is preferred. Smooth sequences
warrant a long short-term memory to take into account far earlier time
points. It usually needs a larger scaling parameter.

Figure \ref{fig_spectralradius} presents the result of the generalization
error rates varying with different scalings. The upside row is for
oscillating series and the downside row is for smooth series. It is clear
from Figure \ref{fig_spectralradius} that severely oscillating sequences
usually require a small spectral scaling, while sequences with slow dynamics
need a larger scaling. However, GunPoint dataset and ShapeletSim dataset
show contrary result. For example, on Gun-Point dataset, which contains
smooth sequences, yet the classification result is optimal when the scaling
is small. ShapeletSim is a dataset which contains oscillating sequences but warrants a large scaling.

\begin{figure*}[!t]
\centering
\includegraphics[width=6.5in]{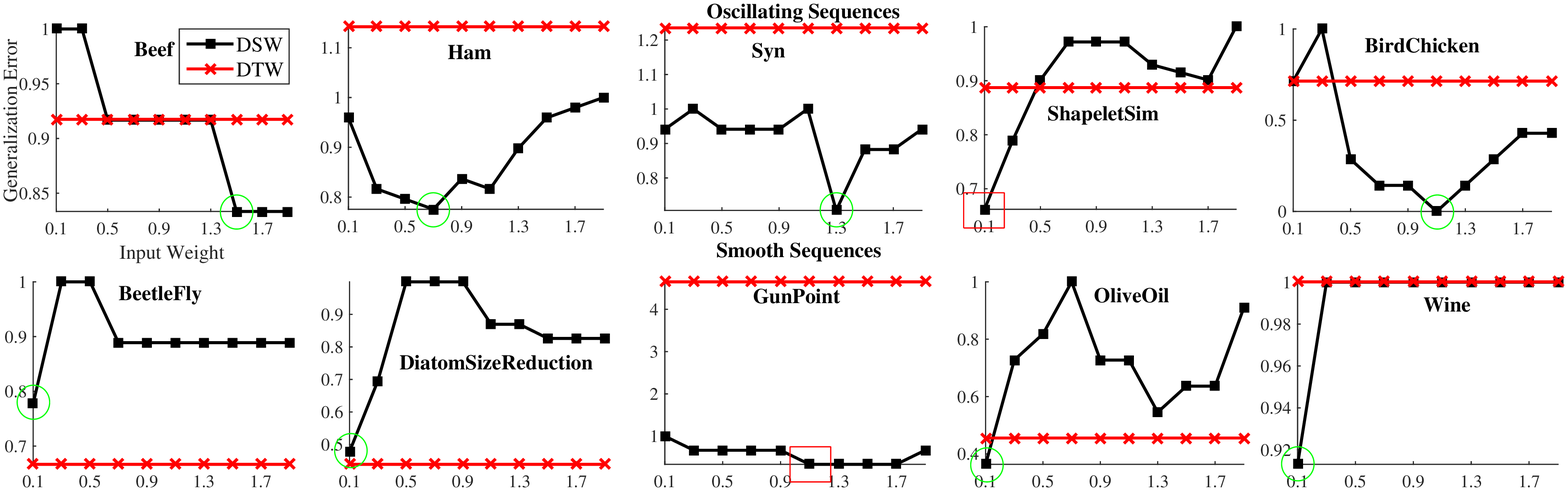}
\caption{Generalization error with varying input weights.
The original error rates on each dataset
have been divided by the maximum error rate for better visualization.
The top row
shows the result of fast oscillating sequences and the bottom row shows the
result of the smooth sequences.
It demonstrates that (1) oscillating sequences usually need a large input weight and
smooth sequences usually need a small input weight (see green circle); (2) local discriminative features need a small input weight and global discriminative features need a large input weight (see red rectangle). In addition, we plot the
DTW generalization error as a comparison. Note that DSW's parameters are set
arbitrarily without optimized by cross-validation. Despite the pessimistic setup of DSW, on most datasets DSW maintains lower error than DTW.}
\label{fig_inputweight}
\end{figure*}

%We have assumed the discriminative characteristics of oscillating sequences
%is mostly local structures while the discriminative features of smoothly changing sequences are global
%structures.
Two factors are responsible for the observation on GunPoint and ShapeletSim datasets:

(1)The first is from the aspect of modeling the sequence dynamics. Generally
speaking, on the one hand, to capture the fast dynamics of sequences, the
reservoir should put less stress on the far earlier time points and
concentrate more on the information of nearby time points. This leads to a
need for a short short-term memory ability. On the other hand, to capture the
difference between smooth sequences, long short-term memory ability is helpful to
capture the global structure information.

(2) The second is the intra-class discriminative features. Since out task at
hand is classification, sometimes the discriminative features are more
important than dynamics modeling. Depending on the discriminative features
are local or global characteristics, a corresponding short-term memory
ability and scaling are needed. In particular, small scaling is helpful if
the discriminative features spread locally. When the discriminative features
are global characteristics, large scaling is preferred.

%On most datasets in our experiment, the same tendency is observed. However,
%on some datasets it is not.
We examine closely on GunPoint dataset. Figure \ref{fig_gunexplain} explains
the observation. The GunPoint dataset contains two classes that differ from
each other by a small region, i.e. taking a gun or no gun, as introduced in
Section \ref{section_introduction}.
The reservoir with a small scaling memory nearby points when converting time points into states.
Therefore, the discriminative features are discovered. According to
Figure \ref{fig_gunexplain}, for a small scaling parameter, the reservoir
provides high dynamics enlarging the minor difference. On the other hand,
when the scaling parameter is large, the reservoir representation is more
smooth, hiding the small difference between two classes. To make the
experiment reproducible, we have fixed the reservoir parameter as above
mentioned.

%ShapeletSim is another example that our prediction of scaling fails.
%However, this is not surprising.
%ShapeletSim is a dataset which contains oscillating sequences but warrants a
%large scaling. In particular,
ShapeletSim dataset contains two class of sequences, which
oscillate severely. The two classes are generated with two frequencies
respectively, which is the main discriminative feature. The within-class
sequences differ from each other by a phase shift, which results in their
different shapes. To capture the inter-class difference, DSW needs long
short-term memory to distinguish the two frequencies, thus the result
(Figure \ref{fig_spectralradius}) shows it performs best when the scaling is
relatively large. %This dataset yields
%challenges for shape based distance algorithms. DTW aligns the sequences and
%counts heavily on the shapes. In this case, it may be fooled by the dataset.
%DSW has a fading memory ability and can generalize better than DTW.

We find two main results in this experiment: (1) Large scaling contributes
to long short-term memory and small scaling contributes to short short-term
memory. In DSW, this can be a guideline in designing reservoir for modeling
sequences. (2) When the discriminative features concentrate on a small
region of sequences, small scaling is preferred to strength them; when the
discriminative features are global structures, large scaling is useful.

To summarize, for classification, the key concern is to discover the
inter-class discriminative features. Normally, despite the specific warrants
of some datasets, on most datasets, it is observed that fast oscillating
dynamics benefit from small scaling parameter and more smooth sequences need
larger scaling parameter. %In particular, notice that the
%general strategy we mentioned above is built on learning faith representations of original
%data, but this result may not match the need of classification.

\subsubsection{Input Weight}

\label{subsub_inputweights}

The reservoir receives the current input and transforms into a new state by
incorporating this input with the previous reservoir state. There is a
trade-off between the current input and previous input history.

We fix the reservoir parameters as the same for the previous subsection. The
input weight varies from 0.1 to 1.9 step by 0.2. Figure \ref{fig_inputweight}
demonstrates the results. It shows small input weights is helpful for long
short-term memory and large input weights for short short-term memory. This
observation is consistent with the result of spectral scaling.

\subsubsection{The Predictability of Reservoir Model}

\label{subsub_predicablity}

\begin{figure}[!t]
\centering
\includegraphics[width=3.3in]{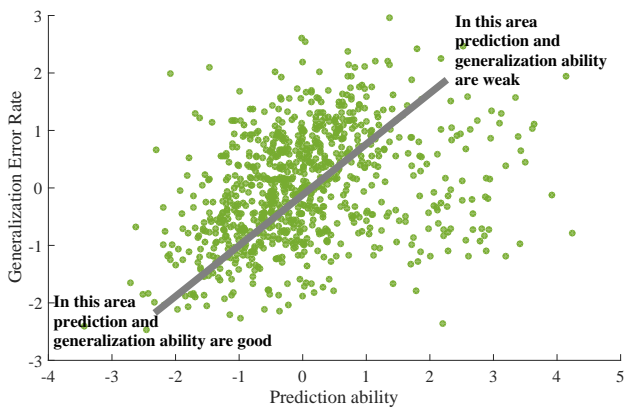}
\caption{The correlation between predictability of reservoir network and the
generalization error rate of DSW. For visualization, the predictability and generalization error are normalized to have zero mean and unitary standard deviation.}
\label{fig_prediction_dsw}
\end{figure}

We employ the predictability of readout layer as a proxy for how well
the state space representation approximate the original sequences.

The reservoir model is trained to approximate the function mapping from
input sequences to output sequences. Define the predictability of a
reservoir as the difference between the empirical output and true output. To
provide more insights into the reservoir model in DSW, we analyze the
relationship between the predictability on training set and classification
performance on test set. We train the reservoir model using one-step forward
prediction. Ridge regression is employed to learn the output weights \cite%
{lukovsevivcius2009reservoir,chen2014learning}. The ridge regression
parameter is selected in $\{10^{-5},10^{-4},\cdots,10 \}$ by 5-fold cross
validation \cite{chen2014learning}. In detail, we run 10 repetitions for
each of the 85 UCR datasets. The reservoir network is regenerated randomly
every time.
%have performed 20 runs of experiments on each dataset in UCR time
%series archive and
The predictability and generalization error of every network is recorded. We
then normalize the two values to have zero mean and unit standard
deviation.
%Then we collect predictability and classification performance of the networks.

By doing so, we can use the predictability of the network on training set as
an indication of the classification performance on test set. We examine the
relationship between the predictability on the training set and the
classification error rate on test set. Figure \ref{fig_prediction_dsw} plots
the correlation between these two values on all datasets. The Pearson
correlation coefficient is 0.3342, which indicates these two variables are
indeed correlated. Therefore, for DSW, it is important to obtain good
classification performance by having a reservoir network that can
approximate the original data well.

\subsubsection{The Size of Dynamic Reservoir}

\label{subsub_reservoirsize}
\begin{figure}[!t]
\centering
\includegraphics[width=3.3in]{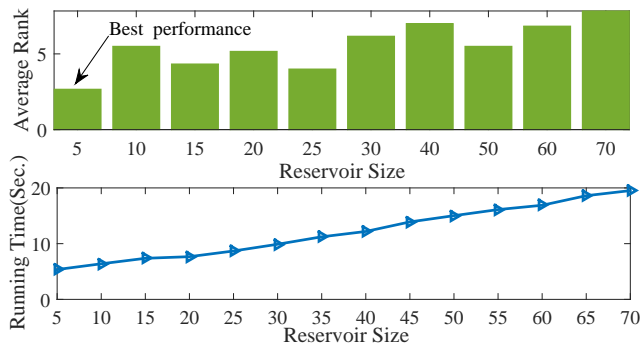}
\caption{The average rank on 42 old UCR time series datasets with respect to
reservoir size (upside). It demonstrates that small reservoir size is enough to yield
good classification results for DSW. And the bottom row presents the running time of DSW w.r.t. reservoir size.}
%The running time for ED and DTW are also plotted for comparison.
%The result indicates that DSW is able to improve the classification performance with tolerable computational cost.}
\label{fig_ReservoirSize}
\end{figure}

The number of neurons in the reservoir has an influence on the memory
capacity of reservoir models \cite{lukovsevivcius2009reservoir}. Large
reservoir provides more nonlinearity and dynamics. To study how the
reservoir size affects DSW, we do experiments on 42 old UCR time series
datasets with different reservoir size, ranging
in\{5,10,15,20,25,30,40,50,60,70\}. We fix the reservoir parameter as $%
r_{i}=0.2$, $r_{c}=0.3$, $r_{j}=0.4$, $N=5$ and $jump length = 2$. On each
dataset, we obtain a set of generalization error rates corresponding to
different reservoir size. The error rates are then sorted so that each value
associates to a rank. The algorithm with the minimum generalization error
obtains the rank of 1 and the second minimum gets the rank of 2 etc.
%The performance rank of each case on a
%dataset is the position of the case in the sorted performance measurement.
The lower rank indicates better performance. We record the average rank of
classification performance on each reservoir size.

Figure \ref{fig_ReservoirSize} presents the experimental result. The upside
row is the average rank for different reservoir size. Surprisingly, it shows
that the reservoir size has very limited influence on the performance of
DSW. The rank of the error rate is good when the reservoir size is relatively
small. In particular, it performs very well when the reservoir size is $5$. It is reasonable to observe this result, since
our task is to discriminate the examples in different classes instead of
modeling the nonlinear dynamics.
The bottom row presents the computational time on Beef dataset for different
reservoir size. We observe that with larger reservoir, more
computational time is need.
DSW is able to improve the classification performance with tolerable computational cost.
%DSW has the same level of time complexity with DTW.

Note that in previous literatures \cite{rodan2012simple,lukovsevivcius2009reservoir}, the size of reservoir is
usually set as hundreds of neurons to capture the generating mechanism. In
our study, we concern different aspects of sequences. The reservoir in our
work is mainly to provide versatile discriminative features to help
classification. As a result, we do not need too large reservoir size.

\subsubsection{Input Dimensionality}

\label{subsubInputDim}
\begin{figure}[!t]
\centering
\includegraphics[width=3in]{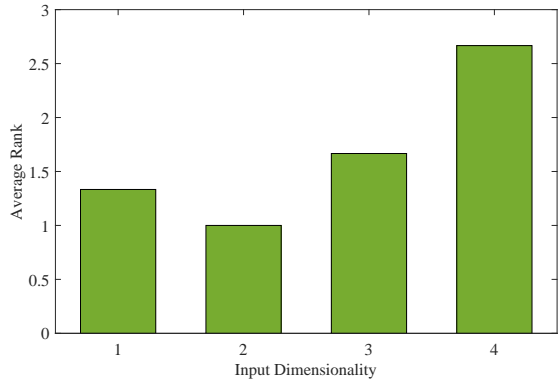}
\caption{The average rank of generalization error for different input
dimensionality.}
\label{fig_InputDim}
\end{figure}
The dimensionality of input influences the resulting state sequences. For $n$
input dimensionality, it means we feed $n$ successive time points as the input into the
reservoir to obtain an updated state. We extend the original sequence by
repeating the end of sequences $n$ times so that the state sequences are of
the same length as the original sequences.

We perform experiments on 42 old UCR datasets using different input
dimensionality. On each dataset, the input dimensionality is selected as
1,2,3 and 4. Then using the selected input dimensionality, we learn
reservoir state sequences. %The experiments are repeated 10 times.
We have fixed the reservoir parameters during the experiment to guarantee
only the input dimensionality is varied. The parameters are the same as that
of subsection \ref{subsub_spectral}. Figure \ref{fig_InputDim} illustrates
the average rank of each dimensionality over all 42 datasets. It clearly
demonstrates that 2 input dimensionality achieves the best performance,
followed by 1 dimensionality. However, for 3 and 4 dimensionality it
performs much poorer.

\section{Conclusion}
In this paper, we propose a novel algorithm, DSW, for calculating the distance between
sequences. DSW employs a reservoir network as a general purpose nonlinear
temporal filter. The original sequences are first converted into reservoir
state trajectory sequences to capture the autocorrelation structure
information. The state trajectory sequence provides versatile discriminative
features for classification. Then dynamic programming is performed to find
an alignment between two state sequences. Therefore, points with
similar states are matched. The time complexity of DSW is at the same level as that of DTW ($O(L_{Q}L_{C})$).

We have conducted extensive experiments to evaluate DSW using both synthetic
datasets and benchmark datasets, compared with DTW and its variants. The experimental results demonstrate the
competitiveness of DSW. In particular, DSW achieves state-of-the-art
classification performance on 85 UCR time series data. DSW is also
empirically demonstrated to be endowed with better robustness and
scalability.

Possible extensions of this work include: (1) to provide more efficient
optimization methods for the reservoir network in DSW. (2) to design
accelerating techniques for searching alignments in DSW.

\bibliographystyle{ieeetr}
\bibliography{DTSWBIB}

\end{document}